\documentclass[runningheads]{llncs}
\usepackage[T1]{fontenc}
\usepackage{xcolor}
\usepackage{booktabs}
\usepackage{multirow}
\usepackage{pifont}
\usepackage{subcaption}
\usepackage{amssymb} 
\usepackage{csquotes}
\usepackage[hidelinks]{hyperref}
\usepackage[flushleft]{threeparttable}

\usepackage{amsmath}
\usepackage{graphicx}
\begin{document}
\title{A Novel Tracking Framework for Devices in X-ray Leveraging Supplementary Cue-Driven Self-Supervised Features}
\titlerunning{SSL with Supplementary Cues and Historical Feature Guided Tracker}

\author{Saahil Islam\inst{1,2}\orcidID{0000-0003-2631-8765} \and 
Venkatesh N. Murthy\inst{3} \and
Dominik Neumann\inst{2} \and 
Serkan Cimen\inst{3} \and
Puneet Sharma\inst{3} \and
Andreas Maier\inst{1} \and
Dorin Comaniciu\inst{3} \and
Florin C. Ghesu\inst{2}
}

% index{Islam, Saahil} 
% index{Murthy, Venkatesh N.}
% index{Neumann, Dominik}
% index{Cimen, Serkan}
% index{Sharma, Puneet}
% index{Maier, Andreas}
% index{Comaniciu, Dorin}
% index{Ghesu, Florin C.}

\authorrunning{S. Islam et al.}

\institute{Friedrich-Alexander-Universität, Pattern Recognition Lab, Erlangen, Germany \and
Digital Technology and Innovation, Siemens Healthineers, Erlangen, Germany \and
Digital Technology and Innovation, Siemens Healthineers, Princeton, NJ, USA\\
\email{saahil.islam@fau.de}}
\maketitle              % typeset the header of the contribution
\begin{abstract}

To restore proper blood flow in blocked coronary arteries via angioplasty procedure, accurate placement of devices such as catheters, balloons, and stents under live fluoroscopy or diagnostic angiography is crucial. Identified balloon markers help in enhancing stent visibility in X-ray sequences, while the catheter tip aids in precise navigation and co-registering vessel structures, reducing the need for contrast in angiography. However, accurate detection of these devices in interventional X-ray sequences faces significant challenges, particularly due to occlusions from contrasted vessels and other devices and distractions from surrounding, resulting in the failure to track such small objects. While most tracking methods rely on spatial correlation of past and current appearance, they often lack strong motion comprehension essential for navigating through these challenging conditions, and fail to effectively detect multiple instances in the scene. To overcome these limitations, we propose a self-supervised learning approach that enhances its spatio-temporal understanding by incorporating supplementary cues and learning across multiple representation spaces on a large dataset.  Followed by that, we introduce a generic real-time tracking framework that effectively leverages the pretrained spatio-temporal network and also takes the historical appearance and trajectory data into account. This results in enhanced localization of multiple instances of device landmarks. Our method outperforms state-of-the-art methods in interventional X-ray device tracking, especially stability and robustness, achieving an 87\% reduction in max error for balloon marker detection and a 61\% reduction in max error for catheter tip detection.

\end{abstract}
\section{Introduction}
\begin{figure}[t]
\centering
\includegraphics[width=\textwidth]{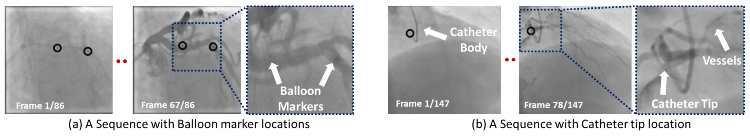}
\caption{Example of (a) balloon markers and (b) catheter tip highlighted in black: indicating the change in appearance over time with the contrast flowing through the vessels.} 
\label{fig:example_case}
\end{figure}

%Application Intro
A clear and stable visualization of the stent is crucial for coronary interventions. Stent enhancement is highly valuable specifically for estimating stent position for under-expansion, stent failure, stent disruption and treatment of aorto-ostial and bifurcation lesions \cite{figini2019use}. Tracked balloon markers can be used as anchor points to stabilize consecutive sequence images and superimposing them to enhance the stent visualization \cite{huang2022robust}. Tracking the catheter tip serves as an anchor point for mapping vessel information between fluoroscopy and angiography images, reducing contrast usage for vessel visualization \cite{ma2020dynamic} and aiding stent and balloon placement in catheterized interventions. Tracking such small objects poses challenges due to complex scenes caused by contrasted vessel structures amid additional occlusions from other devices and from noise in low-dose imaging. Distractions from visually similar image parts along with the cardiac, respiratory and the device motion itself aggravate these challenges. An example of how contrasted vessel structure cause occlusions is depicted in Fig.~\ref{fig:example_case}.

In recent years, various tracking approaches have emerged for both natural and X-ray images. Most of these methods use siamese architectures to extract features from two different crops, typically one search and one or more template frames, enabling them to adapt to changes in appearance via spatial correlation techniques \cite{fan2021cract,yu2020deformable,zhang2021learn,bromley1993signature,li2018high}. Recently, transformers have been integrated into these architectures \cite{yan2021learning,cui2022mixformer,demoustier2023contrack}. However, these methods rely on asymmetrical cropping, which removes natural motion. The small crops are updated based on past predictions, making them highly vulnerable to noise and risk incorrect field of view while detecting more than one object instance. Furthermore, using the initial template frame without an update makes them highly reliant on initialization. To incorporate motion information, some methods use historical trajectory \cite{wei2023autoregressive,lin2022swintrack} which is suboptimal for medical datasets with sparse annotations. Semi-supervised methods like Cycle Ynet \cite{lin2020cycle} have been employed to address the lack of annotated frames, but they may introduce noise due to weak-label supervision. ConTrack \cite{demoustier2023contrack} uses optical flow to integrate contextual spatio-temporal information from past frames, yet it is restricted to a single past frame and relies on mask segmentation, which may not be available for many datasets. Furthermore, self-supervised learning (SSL) approaches have gained popularity by demonstrating how pretraining on unlabeled datasets can enhance performance in downstream tasks \cite{feichtenhofer2021large,qian2021spatiotemporal,feichtenhofer2022masked,tong2022videomae}. FIMAE \cite{fimae} employs a masked image modeling (MIM) based SSL method on a large unlabeled angiography dataset, but it emphasizes reconstruction without distinguishing objects. It's worth noting that the catheter body occupies less than 1\% of the frame's area, while vessel structures cover about 8\% during sufficient contrast. While effective in reducing redundancy, FIMAE's high masking ratio may overlook important local features and focusing solely on pixel-space reconstruction can limit the network's ability to learn features across different representation spaces. Although SimST \cite{fimae} uses a pretrained spatio-temporal encoder for tracking, it still relies on asymmetrical cropping, which may be inefficient for the reasons mentioned above.

In this work, we address the mentioned challenges and improve on the shortcomings of prior methods. The proposed self-supervised learning method integrates an additional representation space alongside pixel reconstruction, through supplementary cues obtained by learning vessel structures (see Fig.~\ref{fig:pipeline}(a)). We accomplish this by first training a vessel segmentation (\enquote{vesselness}) model and generating weak vesselness labels for the unlabeled dataset. Then, we use an additional decoder to learn vesselness via weak-label supervision.
A novel tracking framework is then introduced based on two principles: Firstly, symmetrical crops, which include background to preserve natural motion, that are crucial for leveraging the pretrained spatio-temporal encoder. Secondly, background removal for spatial correlation, in conjunction with historical trajectory, is applied solely on motion-preserved features to enable precise pixel-level prediction. We achieve this by using cross-attention of spatio-temporal features with target specific feature crops and embedded trajectory coordinates.

Our contributions are as follows: % Improve wording
1) Enhanced Self-Supervised Learning using a specialized model via weak label supervision that is trained on a large unlabeled dataset of 16 million frames. 
2) We propose a real-time generic tracker that can effectively handle multiple instances and various occlusions.
3) To the best of our knowledge, this is the first unified framework to effectively leverage spatio-temporal self-supervised features for both single and multiple instances of object tracking applications.
4) Through numerical experiments, we demonstrate that our method surpasses other state-of-the-art tracking methods in robustness and stability, significantly reducing failures.

\section{Methods}
\begin{figure}[ht]
\centering
\includegraphics[width=0.82\textwidth]{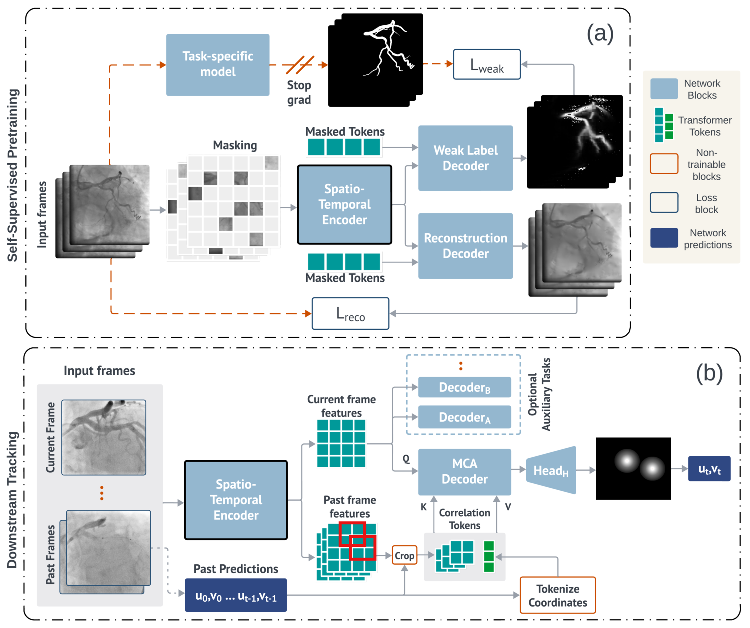}
\caption{Overview of our framework: (a) The Self-Supervised Learning (pretraining) and (b) Historical Guided Tracker leveraging the pretrained features.} 
\label{fig:pipeline}
\end{figure}

Let $\mathcal{D}_u$ denote the large unlabeled dataset and $\mathcal{D}_{s}$ represent a dataset containing pixel-level annotations of vessels. We denote the downstream dataset for tracking as $\mathcal{D}_l$. For the particular objects in consideration, our goal is to track their location, $\hat{y}_t = (u_t, v_t)$ at any time $t, t \geq 0$ given a sequence of X-ray images $\{ I_t\}_{t=0}^\mathcal{N}$  and an initial location $y_0 = (u_0, v_0)$. The proposed self-supervised learning and the downstream tracking framework is depicted in Fig.~\ref{fig:pipeline} and is explained in the subsequent subsections: 

\subsection{Self-Supervised Learning with Supplementary Cues}

We employ a task-specific model to generate weak labels, required for obtaining the supplementary cues. In particular, a U-Net, $\mathcal{F}_{s}(\theta)$, is used to train a \enquote{vesselness} model on $\mathcal{D}_{s}$. The trained model $\mathcal{F}_{s}(\hat{\theta})$ is then utilized to generate vesselness, offline for all sequences $S_k \in \mathcal{D}_u$. For pretraining on the unlabeled dataset $\mathcal{D}_u$, we integrate vesselness supplementary cues into a FIMAE-based MIM model. We denote this as FIMAE-SC for the rest of the manuscript. We sample $n$ frames from $S_k$, $\mathbf{I} \in \mathbb{R}^{n \times h \times w}$ and spatially encode them to $d$ dimensions resulting in $n \times \frac{h}{16} \times \frac{w}{16} \times d$ tokens. The frames are masked with a 75\% tube mask and a 98\% frame mask, followed by joint space-time attention through multi-head attention (MHA) layers. Specifically, each token for the $t^{th}$ frame is projected and flattened into query, key, and value embeddings: $(q_t, k_t, v_t)$, where $t=[0, 1, \ldots , n-1]$ and the joint space-time attention is given by:

\begin{equation}
\label{eqn:attention}
\operatorname{Attention}(\mathrm{Q}, \mathrm{K}, \mathrm{V}) =\operatorname{softmax}\left(\frac{\mathrm{Q}\mathrm{K}^T}{\sqrt{d}}\right) \mathrm{V},
\end{equation}

where variables $(\mathrm{Q}, \mathrm{K}, \mathrm{V})$ are defined as $\mathrm{Q}=\operatorname{Concat}(q_0, q_1, \ldots, q_{n-1}), \
\mathrm{K} = \operatorname{Concat}(k_0, k_1, \ldots, k_{n-1}), \mathrm{V}= \operatorname{Concat} (v_0, v_1, \ldots, v_{n-1})$. The encoded features are projected to a lower dimension $d_{lo}$ and concatenated with two learnable masked tokens (one for each of the subsequent decoders) corresponding to the missing patches resulting in features $f \in \mathbb{R}^{n \times \frac{h}{16} \times \frac{w}{16} \times d_{lo}}$. 
Then, two decoders are employed: a reconstruction decoder, $\mathbf{H}_{R}$ for pixel reconstruction and a weak label decoder, $\mathbf{H}_{W}$ for vesselness prediction, both employing MHA. The respective outputs, $\hat{\mathbf{I}} = \mathbf{H}_{R}(f)$ and $\hat{\mathbf{V}} = \mathbf{H}_{W}(f)$ are projected to $256$ dimensions and reshaped to $n \times h \times w$. The final loss ($\mathcal{L}_u$) is computed as the weighted sum of reconstruction and weak label prediction, $\mathcal{L}_u = \alpha\mathcal{L}_{reco} + \beta\mathcal{L}_{weak}$.

% \begin{equation}
%     \hat{\mathbf{I}} = \operatorname{reshape}(\operatorname{MLP}(\mathbf{H}_{reco}(f_u))
% \end{equation}

% \begin{equation}
%     \hat{\mathbf{V}} = \operatorname{reshape}(\operatorname{MLP}(\mathbf{H}_{weak}(f_u))
% \end{equation}

% \begin{equation}
% \mathcal{L}_{tube}=
% \frac{1}{|\Omega_{T}|}\sum_{t={2\eta}}^n \sum_{p_t \in \Omega_{tube}}\|\mathbf{I}_t(p_t)-\hat{\mathbf{I}}_t(p_t)\|^2 
% \end{equation}

\begin{equation}
\mathcal{L}_{reco}=
\frac{1}{|\Omega_{T}|} \sum_{p \in \Omega_{T}}\|\mathbf{I}(p)-\hat{\mathbf{I}}(p)\|^2 + \frac{|\Omega_{T}|}{|\Omega_{F}|^2}\sum_{q \in \Omega_{F}}\|\mathbf{I}(q)-\hat{\mathbf{I}}(q)\|^2
\end{equation}

% \begin{equation}
% \mathcal{L}_{frame}=\frac{1}{|\Omega_{frame}|}\sum_{t={2\eta+1}}^n \sum_{q_t \in \Omega_{frame}}\|\mathbf{I}_t(q_t)-\hat{\mathbf{I}}_t(q_t)\|^2
% \end{equation}

\begin{equation}
    \mathcal{L}_{weak} = \frac{1}{|\omega|}\sum_{r \in \omega} \|\mathcal{F}_{s}(\hat{\theta})( \mathbf{I}(r))-\hat{\mathbf{V}}(r)\|^2
\end{equation}

Where $p \in \Omega_{T}$ is the token indices of the tube masked tokens, and $\Omega_{T}$ denotes the set of all tube masked token indices. Similarly, $q \in \Omega_{F}$ refers to the frame masked token indices in all randomly frame masked token indices. $\omega$ refers to all tokens. $\alpha$ and $\beta$ are weights assigned for reconstruction and weak label prediction respectively.

\subsection{Historical Feature Guided Tracker}
We design a Historical Feature Guided Tracker (HiFT) for $\mathcal{D}_l$. \\
\textbf{Spatio-temporal encoder.} We input $\hat{n} \in \mathcal{N}$ frames with symmetrical crops to the pretrained spatio-temporal encoder preserving the natural motion. Similar to the pretraining pipeline, each sampled sequence $\hat{n} \times \hat{h} \times \hat{w}$ adopts a joint space-time attention (MHA) to obtain features $\hat{f} \in \mathbb{R}^{\hat{n} \times \frac{\hat{h}}{16} \times \frac{\hat{w}}{16} \times d}$.

% \begin{figure}[ht]
% \centering
% \includegraphics[width=\textwidth]{MSFT (4).png}
% \caption{Overview of our proposed Historical Feature Guided Tracker} 
% \label{fig:msft}
% \end{figure}

\textbf{Dynamic correlation with appearance and trajectory.} We build correlation tokens as a concatenation of appearance and trajectory for modeling relation with past frames. In particular, we use the past frame predictions ($(u_0, v_0), \ldots (u_{{\hat{n}} - 2}, v_{{\hat{n}} - 2})$) as the centre to crop the past frame features $\hat{f}_0, \hat{f}_{1} \ldots \hat{f}_{\hat{n}-2}$ obtaining appearance tokens ($\phi$). To obtain the trajectory ($c$), we tokenize each past frame predicted coordinates similar to SwinTrack \cite{lin2022swintrack} to provide additional information about the motion. We adopt a multi-head Cross-Attention Decoder (MCA) to correlate the current frame features ($\hat{f}_{\hat{n} - 1}$) with the correlation tokens. The output of this decoder is passed through a small Convolutional Neural Network (CNN) head to give a heatmap ($z_{heat}$) corresponding to the locations of the objects to be tracked on the current frame.

\begin{equation}
    z_{heat} = \operatorname{Head_{H}}(\operatorname{MCA}(\hat{f}_{\hat{n} - 1},  Concat(\phi_{0}, c_0, \phi_{1}, c_1, \ldots \phi_{\hat{n}-2}, c_{\hat{n}-2})))
\end{equation}

The coordinates of the landmarks are obtained by grouping the heatmap by connected component analysis (CCA) and obtain argmax (locations) of the number of landmarks (or instances) needed to be tracked. We adopt auxiliary decoders ($Decoder_A, Decoder_B, \ldots$) for datasets $\mathcal{D}_l$, where additional annotations are present, e.g. dense mask annotations of catheter body. An auxiliary decoder simply follows MHA with a task-specific head, predicting $z_{aux}$. We use a weighted loss, $\mathcal{L}_l = \mathcal{L}_P + \sum^{\mathcal{J}}_{j=1} \lambda_j\mathcal{L}_j$ as our loss function for $\mathcal{J}$ auxiliary tasks. $\lambda_{j}$ denotes the weights assigned to each auxiliary task. $\mathcal{L}_P$ follows a soft dice loss given by:  

% \begin{equation}
%     (u_t,v_t) = \operatorname{argmax}_{lnd}(\operatorname{CCA}(z_{heat}))
% \end{equation}

\begin{equation}
\mathcal{L}_P = 1 - \frac{2 * \sum G * z_{heat}}{\sum G^2+\sum z_{heat}^2+\epsilon}
\end{equation}

% \begin{equation}
% \mathcal{L}_M=\left\{\begin{array}{ll}
% \frac{2 * \sum G_{mask} * z_{mask}}{\sum G_{mask}^2+\sum z_{mask}^2+\epsilon}, & \text { if } G_{mask} \text { exists } \\
% 0 & \text { otherwise }
% \end{array},\right.
% \end{equation}

where G represents ground truth labels. We use a similar dice loss for catheter body mask prediction as our auxiliary loss for frames where such annotations are available with $\lambda$ of 0.5.

\section{Experiments}
%\subsection{Dataset and Experimental Setup}
\textbf{Dataset.} The vesselness dataset ($\mathcal{D}_s$) consists of 3300 training and 91 testing angiography sequences. Coronary arteries were annotated with centerline points and approximate vessel radius for 5 sufficiently contrasted frames, which were then used to generate target vesselness maps for training.
The unlabeled dataset ($\mathcal{D}_u$) includes 241,362 sequences from 21,589 patients, totaling 16,342,992 frames, comprising both angiography and fluoroscopy sequences.
We use two downstream datasets ($\mathcal{D}_l$) for evaluating the tracking performance. The balloon marker dataset consists of 1058 training and 113 test sequences consisting of both fluoroscopy and angiography sequences. All frames are annotated with the location of the balloon marker pairs. For the catheter tip dataset, there are 2,314 training sequences totaling 198,993 frames, with annotations for 44,957 frames, and 219 test sequences with complete frame annotations. A subset of the training dataset includes catheter body mask annotations. Both test datasets are divided into \enquote{with occlusion}, where at least one frame in the sequence is obstructed, and \enquote{no occlusion}, where the entire sequence is free of obstruction. The balloon marker dataset has a ratio of 38:75 for \enquote{with occlusion} to \enquote{no occlusion} cases, while the catheter tip dataset has a ratio of 125:94. 

\textbf{Experimental Setup:} We adopt a similar preprocessing pipeline as ConTrack \cite{demoustier2023contrack}. During training, we randomly sample 5 consecutive annotated frames, cropping them to 256x256 using the first frame annotation as the center. During inference, similar crops are applied and updated if the distance from the past prediction to the border exceeds 30 pixels. Please refer to supplementary materials for more details. We train for 250 epochs using a learning rate of 0.0002.

%\subsection{Results and Discussions}
\textbf{Comparison with State-of-the-Art.} We assess our approach's performance against existing methods in Table \ref{tab:allresults} for balloon marker and catheter tip detection with both manual and automatic initialization. Most trackers rely on modeling appearance changes, particularly advantageous for catheter tip tracking, where the tip is often entirely occluded during contrast injection. While these methods demonstrate similar precision in detecting balloon markers, their high standard deviation and max error indicate inadequate motion comprehension, particularly critical for such small objects, which are vulnerable to distractions. 3D-DenseUNet utilizes multiple uncropped frames (as channels) preserving natural motion, leading to comparable performance to the specialized trackers for balloon marker tracking, but fails to track catheter tip due to the absence of modeling for appearance changes. Our approach integrates both advantages, significantly reducing max error by 87\% and 61\% for balloon markers and catheter tip tracking respectively, resulting in highly stable and robust performance. 3D-DenseUNet utilizes multiple full-sized frames, eliminating the need for initialization. To enable automatic initialization for other trackers, we train a detection model with the same backbone as the tracker followed by upsampling conv layers, using a single uncropped frame as input. Predictions from this model serve as the initialization. Due to our approach's robustness and reduced reliance on initialization, it achieves either superior or comparable performance to the prior manually-initialized trackers, even when using automatic initialization.

\begin{table}[t]
\centering
\begin{threeparttable}
\centering
\caption{Performance comparison of different tracking models in terms of average distance (RMSE) in mm. Accuracy improvement for balloon marker and catheter tip is statistically significant with p-value < 0.0005 and p-value <0.05 respectively over the best existing state-of-the-art method for both manual and automatic initialization.}
\label{tab:allresults}
%\resizebox{0.97\columnwidth}{!}{%
\begin{tabular}{c|cccc|cccc|c}
Model  & \multicolumn{4}{c}{RMSE - Balloon Marker} & \multicolumn{4}{c}{RMSE - Catheter Tip}  & FPS            \\ \hline
    & mean & median & std  & max  & mean & median  & std  & max &           \\ \hline
\multicolumn{10}{c}{\textbf{With Manual Initialization}} \\ \hline
 SiameseRPN \cite{li2018high}  & 11.76  & 10.61  & 5.94   &  40.36       & 9.01 & 7.13 & 6.81   & 46.23 &  18   \\
 Mixformer \cite{cui2022mixformer} & 2.32  & 0.64 & 4.49  &  33.43      & 5.15 & 2.68 & 7.10   & 49.29   &  20   \\
 Stark \cite{yan2021learning}       & 1.38  & 0.36  &  3.12  &  27.51      & 4.14 & 2.65 & 4.93   & 31.34  &  22   \\
 Cycle YNet \cite{lin2020cycle} & 1.66  & 0.30  & 4.38   & 22.82        & 2.68  & 1.96 & 2.40   & 21.04   &   109   \\
 ConTrack \cite{demoustier2023contrack}   & 1.37 & 0.32 & 3.08 & 20.25    & 1.63 & 1.08 & 1.70 & 13.32   &  21/12*    \\
 SimST \cite{fimae}      & 1.33 & 0.38 & 2.84 & 23.50    & 1.44 & \textbf{1.02} & 1.35 & 10.23  &   46/42*    \\
 % SimST-FimV (Ours)    & 0.91 & 0.30 & 1.89 & 15.05    & 1.35 & \textbf{0.95} & 1.15 & 9.35   &  46/42*     \\
 % HiFT-Fim (Ours) & 0.45 & 0.28 & 0.61 & 4.82    & 1.33 & 1.08 & 0.81 & 4.99 & 31/28* \\ 
 HiFT (Ours) & \textbf{0.31} & \textbf{0.24} & \textbf{0.28} & \textbf{2.68} & \textbf{1.21} & 1.04 & \textbf{0.68}  & \textbf{4.04} & 31/28* \\ \hline
 
 \multicolumn{10}{c}{\textbf{With Automatic Initialization}} \\ \hline
  3D-DenseUNet \cite{li2018h}   & 1.37 & 0.32 & 2.98  & 21.33     & 9.75  & 7.38   & 7.01    &  53.56        & 87   \\
ConTrack \cite{demoustier2023contrack} &  1.54 &   0.36   &  3.01  &  32.45  & 2.87 &  2.29   & 2.36 & 17.26 & 21/12*  \\
SimST \cite{fimae}      & 1.58    &   0.41      &   2.70     &  38.14      &  2.24     &   1.61      &  2.19      &  18.66 & 46/42*     \\
HiFT (Ours)     &  \textbf{0.57}      &  \textbf{0.24}        & \textbf{1.73}       & \textbf{16.61}       & \textbf{1.45}  & \textbf{1.05}    & \textbf{1.30}  & \textbf{9.29} & 31/28*

\end{tabular}
%}
\begin{tablenotes}
      \small
      \item *catheter tip tracking runtime is slower than balloon markers for trackers with modules dependent on catheter body mask predictions.
    \end{tablenotes}
\end{threeparttable}
\end{table}

\begin{figure}[ht]
	\centering
	\includegraphics[width=\textwidth]{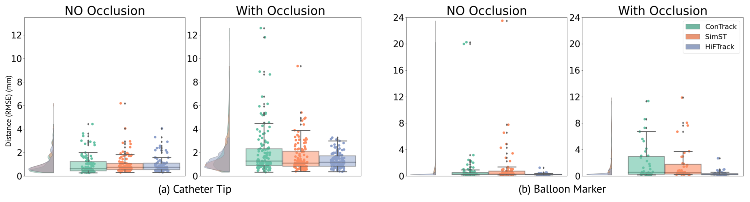}
	\caption{Error distribution for scenarios with and without occlusion for (a) catheter tip and (b) balloon marker tracking} 
 \label{fig:occlusion}
\end{figure}

\begin{figure}[t]
	\centering
	\includegraphics[width=\textwidth]{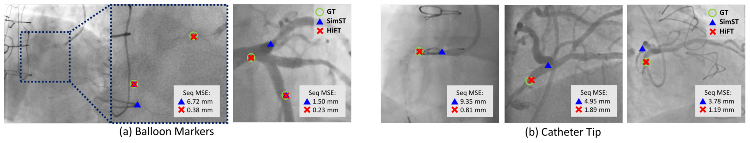}
	\caption{Qualitative examples of balloon marker and catheter tip tracking: Robust performance of HiFT across occlusions and distractors compared to SimST.} 
	\label{fig:qual}
\end{figure}

\textbf{Performance for scenarios with occlusion.} We compare our approach with prior trackers while separating occlusion and no occlusion cases for manual initialization (A similar trend is observed for automatic initialization) in Fig.~\ref{fig:occlusion}. Tracking amid occlusions is challenging for both the applications, leading to greater errors in precision. However, no occlusion cases in balloon marker have a greater number of failures due to their vulnerability to distractors like noise and visually similar image parts. Despite these difficulties, our method demonstrates superior performance across all scenarios. Examples of our method's robust tracking amidst different kinds of scenes is illustrated in Fig. \ref{fig:qual}.

%\begin{table}[ht]
% \centering
% 	\caption{Tracking performance with model predicted initialization.}
% 	\label{tab:noinit}
% 	\begin{tabular}{c|cccc|cccc}
% 		Model        & \multicolumn{4}{c}{balloon markers} & \multicolumn{4}{c}{catheter tip} \\ \hline
% 		& $\mu$ & 95\%P & $\sigma$  & max  & $\mu$ & 95\%P  & $\sigma$  & max    \\ \hline
% 		DenseUnet & 1.37   & 0.32     & 2.98   & 21.33  & 9.75  & 7.38    & 7.01  & 53.56  \\
% 		ConTrack &  1.19  &   0.25   &  2.70  &  20.36  & 2.87 &  2.29   & 2.36 & 17.26  \\
% 		SimST       & 0.95    &   0.30       &   2.42     &  22.74      &  2.24     &   1.61      &  2.19      &  18.66      \\
% 		HiFT (Ours)     &  \textbf{0.57}      &  \textbf{0.24}        & \textbf{1.73}       & \textbf{16.61}       & \textbf{1.45}  & \textbf{1.05}    & \textbf{1.30}  & \textbf{9.29}
% 	\end{tabular}
%\end{table}

\begin{table}[htp]
    \centering
    \begin{minipage}[t]{.66\linewidth}
    \raggedright
	%\centering
	\caption{Comparison of pretraining strategy \\ for tracking}
	\label{tab:pretrain}
	\begin{tabular}[t]{c|c|ccc}
		Pretraining & Downstream & \multicolumn{3}{c}{RMSE} \\ 
		Model & Tracking & mean & std  & max    \\ \hline
		FIMAE & SimST     & 1.44   & 1.35  & 10.23  \\
		FIMAE-SC (Ours) &  SimST     & 1.35  & 1.15  & 9.35  \\
        None & HiFT (Ours) & 3.21 & 2.34 & 16.16 \\
		FIMAE       & HiFT (Ours)        & 1.33  & 0.81  & 4.99  \\
		FIMAE-SC (Ours)  & HiFT (Ours) & \textbf{1.21}  & \textbf{0.68}  & \textbf{4.04}  \\
	\end{tabular}
    \end{minipage}
    \begin{minipage}[t]{.33\linewidth}
   \centering
\caption{Effect of appearance ($\phi$) and trajectory ($c$) tokens for HiFT}
\begin{tabular}{c|c|ccc}
$\phi$ & $c$ & mean & std & max \\ \hline
\ding{55} & \ding{55} & 2.13     & 2.01     & 19.49    \\
\ding{55} & \ding{51} & 1.73    & 1.29    &  12.04   \\
\ding{51} & \ding{55} & 1.40     & 0.88    &  6.41   \\
\ding{51} & \ding{51}& \textbf{1.21}   &  \textbf{0.68} &   \textbf{4.04}  \\  
\end{tabular}
\label{tab:correlation}
    \end{minipage}
\end{table}

\textbf{Ablations.}
We perform ablations on catheter tip tracking with manual initialization to compare pretraining strategies in Table \ref{tab:pretrain}. We observe that the proposed SSL with supplementary cues has a clear advantage irrespective of the downstream tracking framework, whereas the tracking performance without any pretraining drops significantly. The effect of attending to the correlation tokens, i.e., appearance and the trajectory, is explored in Table \ref{tab:correlation}. The best results are obtained when both correlation tokens are attended at the MCA decoder. While trajectory aids performance, the primary enhancement can be attributed to appearance tokens due to sparse annotations in the dataset, constraining the network to fully understand the trajectory during training.

\section{Conclusion}
In this work, we enhance Self-Supervised Learning by incorporating contextual cues through weak-label supervision, encouraging the network to learn features across multiple representation spaces. We introduce a novel tracking framework leveraging the pretrained spatio-temporal network for device tracking, substantially reducing failures compared to prior state-of-the-art methods. Our approach shows promising results even without manual initialization. As a future work, the self-supervised learning method encourages us to explore more than 2 representation spaces and use the pretrained network for tasks other than tracking. While we use a naive method to test our performance without manual initialization, automatic initialization based tracking requires further investigation.

\begin{credits}

\subsubsection{\discintname}
The authors have no competing interests to declare that are
relevant to the content of this article.
\end{credits}


\bibliographystyle{splncs04}
\bibliography{references}

\begin{thebibliography}{10}
\providecommand{\url}[1]{\texttt{#1}}
\providecommand{\urlprefix}{URL }
\providecommand{\doi}[1]{https://doi.org/#1}

\bibitem{bromley1993signature}
Bromley, J., Guyon, I., LeCun, Y., S{\"a}ckinger, E., Shah, R.: Signature
  verification using a" siamese" time delay neural network. Advances in neural
  information processing systems  \textbf{6} (1993)

\bibitem{cui2022mixformer}
Cui, Y., Jiang, C., Wang, L., Wu, G.: Mixformer: end-to-end tracking with
  iterative mixed attention. In: Proceedings of the IEEE/CVF Conference on
  Computer Vision and Pattern Recognition. pp. 13608--13618 (2022)

\bibitem{demoustier2023contrack}
Demoustier, M., Zhang, Y., Murthy, V.N., Ghesu, F.C., Comaniciu, D.: Contrack:
  contextual transformer for device tracking in x-ray. arXiv preprint
  arXiv:2307.07541  (2023)

\bibitem{fan2021cract}
Fan, H., Ling, H.: Cract: Cascaded regression-align-classification for robust
  tracking. In: 2021 IEEE/RSJ International Conference on Intelligent Robots
  and Systems (IROS). pp. 7013--7020. IEEE (2021)

\bibitem{feichtenhofer2021large}
Feichtenhofer, C., Fan, H., Xiong, B., Girshick, R., He, K.: A large-scale
  study on unsupervised spatiotemporal representation learning. In: Proceedings
  of the IEEE/CVF Conference on Computer Vision and Pattern Recognition. pp.
  3299--3309 (2021)

\bibitem{feichtenhofer2022masked}
Feichtenhofer, C., Li, Y., He, K., et~al.: Masked autoencoders as
  spatiotemporal learners. Advances in neural information processing systems
  \textbf{35},  35946--35958 (2022)

\bibitem{figini2019use}
Figini, F., Louvard, Y., Sheiban, I.: Use of stent enhancement technique during
  percutaneous coronary intervention--a case series. Heart International
  \textbf{13}(1), ~28 (2019)

\bibitem{huang2022robust}
Huang, L., Liu, Y., Chen, L., Chen, E.Z., Chen, X., Sun, S.: Robust
  landmark-based stent tracking in x-ray fluoroscopy. In: European Conference
  on Computer Vision. pp. 201--216. Springer (2022)

\bibitem{fimae}
Islam, S., Murthy, V.N., Neumann, D., Das, B.K., Sharma, P., Maier, A.,
  Comaniciu, D., Ghesu, F.C.: {Self-supervised learning for interventional
  image analytics: toward robust device trackers}. Journal of Medical Imaging
  \textbf{11}(3),  035001 (2024). \doi{10.1117/1.JMI.11.3.035001}

\bibitem{li2018high}
Li, B., Yan, J., Wu, W., Zhu, Z., Hu, X.: High performance visual tracking with
  siamese region proposal network. In: Proceedings of the IEEE conference on
  computer vision and pattern recognition. pp. 8971--8980 (2018)

\bibitem{li2018h}
Li, X., Chen, H., Qi, X., Dou, Q., Fu, C.W., Heng, P.A.: H-denseunet: hybrid
  densely connected unet for liver and tumor segmentation from ct volumes. IEEE
  transactions on medical imaging  \textbf{37}(12),  2663--2674 (2018)

\bibitem{lin2020cycle}
Lin, J., Zhang, Y., Amadou, A.a., Voigt, I., Mansi, T., Liao, R.: Cycle ynet:
  semi-supervised tracking of 3d anatomical landmarks. In: Machine Learning in
  Medical Imaging: 11th International Workshop, MLMI 2020, Held in Conjunction
  with MICCAI 2020, Lima, Peru, October 4, 2020, Proceedings 11. pp. 593--602.
  Springer (2020)

\bibitem{lin2022swintrack}
Lin, L., Fan, H., Zhang, Z., Xu, Y., Ling, H.: Swintrack: a simple and strong
  baseline for transformer tracking. Advances in Neural Information Processing
  Systems  \textbf{35},  16743--16754 (2022)

\bibitem{ma2020dynamic}
Ma, H., Smal, I., Daemen, J., van Walsum, T.: Dynamic coronary roadmapping via
  catheter tip tracking in x-ray fluoroscopy with deep learning based bayesian
  filtering. Medical image analysis  \textbf{61},  101634 (2020)

\bibitem{qian2021spatiotemporal}
Qian, R., Meng, T., Gong, B., Yang, M.H., Wang, H., Belongie, S., Cui, Y.:
  Spatiotemporal contrastive video representation learning. In: Proceedings of
  the IEEE/CVF Conference on Computer Vision and Pattern Recognition. pp.
  6964--6974 (2021)

\bibitem{tong2022videomae}
Tong, Z., Song, Y., Wang, J., Wang, L.: Videomae: masked autoencoders are
  data-efficient learners for self-supervised video pre-training. Advances in
  neural information processing systems  \textbf{35},  10078--10093 (2022)

\bibitem{wei2023autoregressive}
Wei, X., Bai, Y., Zheng, Y., Shi, D., Gong, Y.: Autoregressive visual tracking.
  In: Proceedings of the IEEE/CVF Conference on Computer Vision and Pattern
  Recognition. pp. 9697--9706 (2023)

\bibitem{yan2021learning}
Yan, B., Peng, H., Fu, J., Wang, D., Lu, H.: Learning spatio-temporal
  transformer for visual tracking. In: Proceedings of the IEEE/CVF
  international conference on computer vision. pp. 10448--10457 (2021)

\bibitem{yu2020deformable}
Yu, Y., Xiong, Y., Huang, W., Scott, M.R.: Deformable siamese attention
  networks for visual object tracking. In: Proceedings of the IEEE/CVF
  conference on computer vision and pattern recognition. pp. 6728--6737 (2020)

\bibitem{zhang2021learn}
Zhang, Z., Liu, Y., Wang, X., Li, B., Hu, W.: Learn to match: automatic
  matching network design for visual tracking. In: Proceedings of the IEEE/CVF
  International Conference on Computer Vision. pp. 13339--13348 (2021)

\end{thebibliography}

\end{document}